\documentclass[twocolumn]{article}

\usepackage{flushend}
\usepackage{amssymb}
\usepackage{amsmath}
\usepackage{graphicx}
\usepackage{dcolumn}
\usepackage[hidelinks,colorlinks=true,linkcolor=blue,citecolor=blue]{hyperref}
\usepackage{titlesec}
\usepackage{apacite}
\usepackage{authblk}
\usepackage{natbib}
\usepackage{booktabs}
\usepackage[letterpaper, margin=1in]{geometry}
\begin{document}

\title{Fine-Grained Emotional Paraphrasing along Emotion Gradients}
\author{Justin J. Xie}
\affil{Westview High School \\ Portland, OR 97229 \\ justinjxie@gmail.com}
\date{}

\maketitle  
\begin{abstract}
Paraphrase generation, a.k.a. paraphrasing, is a common and important task in natural
language processing. Emotional paraphrasing, which changes the emotion embodied in a 
piece of text while preserving its meaning, has many potential applications, e.g., 
moderating online dialogues and preventing cyberbullying. We introduce a new task of fine-grained emotional 
paraphrasing along emotion gradients, that is, altering the emotional intensities 
of the paraphrases in fine grain following smooth variations in affective dimensions 
while preserving the meanings of the originals. We propose a framework for addressing 
this task by fine-tuning text-to-text Transformers through multi-task training. We 
enhance several widely used paraphrasing corpus by annotating the input and target 
texts with their fine-grained emotion labels. With these labels, fine-tuning text-to-text 
Transformers on these corpus entails multi-task training. Evaluations of the fine-tuned 
Transformers on separate test sets show that including fine-grained emotion labels in 
the paraphrase task significantly improve the chance of obtaining high-quality paraphrases of the desired emotions, i.e., more than doubling the number of exact matches of desired emotions while achieving consistently better scores in paraphrase metrics such as BLEU, ROGUE, and METEOR.  
\end{abstract}

\section{Introduction}
With the rise of social media and online chat rooms, 
the textual aspect of language is often found to be the only aspect of communication that is transferred over the Web. 
Devoid of any intonations or accompanying facial movements, 
it is more challenging for people to decipher the true meaning of messages, 
especially if that message incorporates the more complex aspects of speech (e.g. sarcasm).
Furthermore, 
people often find themselves sending messages without reviewing the message carefully.
For example, political tweets from prominent figures without a careful consideration can lead to political radicalization and conflicts. 
On messaging apps such as Discord, cyberbullies attack others by emotion-ladened words while potentially innocent people unintentionally send unnecessarily intense emotional messages in the heat of the moment. 
Emotional moderation could be an important solution to the overt and extreme emotions expressed on social media.

Paraphrase generation (a.k.a. paraphrasing) is a key task in the domain of natural language processing.
It involves generating an output text that preserves the meanings of the input text while including  
variations in words and grammars. 
The refined task of emotional paraphrasing has garnered much attention. 
Its goal is to alter the underlying emotion associated with the target sentence while maintaining its meaning. 

In this paper, we introduce a new task of fine-grained emotional paraphrasing along emotion gradients. 
This task involves altering the sentiment of a text (e.g. from anger to annoyance) following smooth 
variations in affective dimensions while preserving the meaning. 
It has many potential applications, e.g., moderating online dialogues and preventing cyberbullying. 
We have developed a framework for tackling this task by fine-tuning text-to-text Transformers through
multi-task training. We propose the concept of an emotion-transition graph, whose transitions are based 
on the fine-grained emotion categories and their emotion gradients as identified by  
GoEmotions~\citep{demszky2020goemotions}, and are further curated manually to be more reflective of natural dialogues.  
Our framework works as follows: 
\begin{itemize} 
\item \textbf{Emotion-Labeling of Training Datasets:} Given the paraphrase datasets, we apply a fine-grained 
emotion classification model to label the source and target of each paraphrase pair with their emotions. 
\item \textbf{Fine-Tuning of Text-to-Text Transformer Models:} Utilizing the emotion-labeled training and test datasets, 
we fine-tune a selected text-to-text Transformer model by adding the emotion transition: ``(source emotion) to 
(target emotion)'' as the prefix to the source text. 
\item \textbf{Paraphrasing with Emotional Transition Graph:} Given an input text to be paraphrased, we first obtain
its emotion label and use it identify suggested target emotions through the emotion transition
graph. We then use the fine-tuned text-to-text transformer model to conduct inference on the input text with 
the prefix: source emotion to target emotion. 
\end{itemize} 

In the prototype implementation of our framework, we adopted GoEmotions as our fine-grained emotion
classification model. The GoEmotions model is a multi-labeler, i.e., it provides a list of possible
emotions with their ``likelihood" scores. We modified the model to only report the dominant emotion that 
were above a certain threshold. If no emotion met the threshold, the model reported no emotion label. 
We applied the modified GoEmotions model to a combined dataset that included Google, Microsoft, Quora, and
Twitter paraphrasing corpus. As we labeled paraphrase pairs with emotions, we dropped those pairs that missed 
at least one emotion label due to not meeting the threshold. We utilized 
T5~\citep{raffel2020exploring} as our text-to-text Transformer model. The design of T5 as a multi-task learner allows the input 
to have a prefixes to indicate its specific task. For 
emotional paraphrase fine-tuning, we created prefixes in the format: ``(input emotion) to 
(target emotion)" to help guide the T5 model in identifying and executing our specific task.
The emotional transition graph for our prototype was constructed based on the emotion adjacency matrix 
and clustering as proposed in GoEmotions with additional manual adjustment. 

For performance evaluation of our fine-tuned models for fine-grained emotional paraphrasing, 
emotion-transition performance was evaluated by comparing the emotion of the target text with the emotion of the output text and obtaining 
the Exact Match percentage. The paraphrasing performance was evaluated using the BLEU, ROUGE, and METEOR metrics.
To evaluate the transferability of our fine-tuned model and its sensitivities to the sizes and varieties
of training data. We created several training/evaluation configurations such as small vs large amounts of 
training data and separate vs unified datasets. 

In all performance categories, the fine-tuned models 
showed significant performance improvements over the base T5 model on their respective datasets. 
The fine-tuned models are less sensitive to the sizes of training datasets, but are more sensitive to 
the variety of training datasets. This study indicates that this fine-grained emotional paraphrasing 
framework has major potential in applications to specific scenarios such as chat rooms, forums, and 
public online spaces. It can perform well after fine-tuning on limited data from these application scenarios.

\section{Background}

\subsection{Emotion Psychology}
Emotions are a key component of human psychology. 
They play a role in many cognitive processes including learning, memory, decision making, and interpersonal communication \citep{oatley1994experience, tyng2017influences}. 
Emotions dictate the context and mood by which memories are formed. 
They guide our conscious actions and our body’s unconscious processes. 
An emotion like fear signals the Sympathetic Nervous System to constrict blood vessels, dilate the pupils, and increase the heart rate \citep{alshak2019neuroanatomy}. 
Emotions connect the body’s systems and one’s thoughts towards what is triggering the intense flow of emotions \citep{oatley1994experience}.

Equally important is the role that emotions play in human-to-human interactions. 
Small intonation changes and a rise and lowering of the pitch in one's voice can also signal different emotions \citep{tiwari2012voice}. 
The quick registration of emotions by the body’s senses is efficient for communication. However, 
because emotions act as “heuristics” for the human body \citep{oatley1994experience}, 
it is also quick, and often too quick, to offend in its response in communication. 
Words can trigger emotional responses, both negative and positive. 
Words and languages have been shown to play a key role in helping people distinguish the emotions they feel \citep{lindquist2015role}.

In the Internet age, where 72\% of United States adults are on social media ~\citep{pewresearch2022socialmediafactsheet},
the importance of words in communication has grown. 
Without facial expressions, vocal intonations, or hand gestures, 
it is harder to communicate one’s emotions online. 
The intensities of words can be misinterpreted or be higher than what someone wants them to communicate. 
For example, 
someone could want to communicate frustration, but instead could come off as furious. 
Rooted in the psychology of communication and emotion, the need for lowering the intensity of online communications inspires the task of emotion-paraphrasing along emotion gradients.

\subsection{Sentiment Analysis} 
Sentiment analysis is the NLP task of finding the opinion, attitude, or view of a given input text. 
One sub task of sentiment analysis is the polarity classification which involves classifying the text into one of three basic categories: positive, negative, and neutral. This can be done through supervised learning, a lexicon-based approach, or a hybrid of the two \citep{sadia2018overview}. 
Polarity classification has been done on a variety of sentence types such as reviews \citep{turney2002thumbs, pang2002thumbs}, conditional sentences \citep{narayanan2009sentiment}, and Tweets \citep{pak2010twitter}.
However, 
polarity classification does not allow detailed identification of emotions that is necessary for emotional paraphrasing. 
Emotion classification is needed for this task.

\subsection{Emotion Classification}
In 1890, 
William James proposed fear, grief, love, and rage as a set of the most basic emotions \citep{james1890principles}. 
Then in 1992, 
Paul Ekman introduced his famous set of six basic emotions: fear, anger, joy, sadness, disgust, and surprise \citep{ekman1992argument}. 
They are the basis of many emotional psychology studies and NLP experiments pertaining to emotions. 
Another classification produced by \citeauthor{lazarus1994passion} included a list of 15 emotions. 
Recently a study done by \citeauthor{cowen2017self} expanded on these classifications. 
By having human test subjects report on the emotions they felt while viewing videos, 
the study found that there were 27 emotion categories, in addition to a neutral emotion. 
This study also grouped these emotions into ``clusters.''

\citeauthor{demszky2020goemotions} produced a similar set of 28 emotions that was used in the GoEmotions project. 
This project provided a labeled dataset of 58k texts and a BERT-based machine learning model capable of classifying inputs into one of the 28 emotions. 
In addition, 
the GoEmotions project provided a heat map showing the correlation between emotions as well as including a stratification of the emotions into clusters as shown in Table~\ref{EmotionClusters}. 
These clusters provided guidance for how our emotion-transition graph can be structured. 

\begin{table}[htp]
\caption{Emotion Grouping by \citeauthor{demszky2020goemotions}}
\centering
\small
\noindent
\def\arraystretch{1.5}
\begin{tabular}{ c | l }
\hline
\textbf{Group} & \textbf{Emotions} \\
\hline
 1 & Neutral \\
 2 & Amusement, Excitement, Joy, Love \\
 3 & Optimism, Desire, Caring \\
 4 & Pride, Admiration \\
 5 & Gratitude, Relief \\
 6 & Approval, Realization \\
 7 & Surprise, Curiosity, Confusion \\
 8 & Fear, Nervousness \\
 9 & Remorse, Embarrassment \\ 
 10 & Disappointment, Sadness, Grief \\
 11 & Disgust, Anger, Annoyance, Disapproval
\end{tabular}
\label{EmotionClusters} 
\end{table}

\subsection{Transformers \\ in Natural Language Generation} 
Natural Language Generation (NLG) include tasks such as summarization, translation, chatbots, question generation, long text generation, and paraphrase generation \citep{celikyilmaz2020evaluation}. 
One of the breakthrough NLG models is the Transformer model \citep{vaswani2017attention}, which utilizes the mechanism of self-attention: differentially weighting the significance of different parts of the input data. Bidirectional Encoder Representations from Transformers (BERT) is a well-known language model based on Transformers and pretrained for two tasks: language modeling and next sentence generation~\citep{devlin2018bert}. The GoEmotions' emotion classification model is based on BERT. One of the most complex Transformer-based models is Generative Pre-trained Transformer (GPT), an autoregressive language model. Both of its recent editions, GPT-2~\citep{radford2019language} and GPT-3~\citep{brown2020gpt3} were trained with a dataset containing billions of parameters capable of a wide variety of tasks.

\begin{figure*}[t!]
    \centering
    \includegraphics[width=\linewidth]{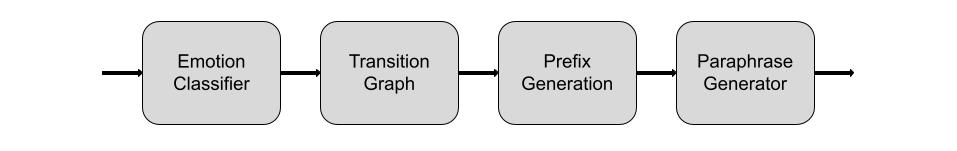}
    \caption{Workflow of Fine-Grained Emotional Paraphrasing along Emotion Gradients}
    \label{fig:Approach1}
\end{figure*}

Text-to-text Transformer models such as Bart \citep{lewis2019bart} and T5  \citep{raffel2020exploring} are Transformer models following the text-to-text approach. They provide a high level of flexibility in NLG tasks because many problems can be solved by a text-to-text Transformer model through similar training or fine-tuning methods. Both Bart and T5 use the standard encoder-decoder Transformer model. 
BART utilizes a technique called ``denoising" for pretraining sequence-to-sequence models based on a combination of Bidirectional and Auto-Regressive Transformers, while
T5 was developed through a process of transfer learning. 
Transfer learning involves training the model on a vast set of wide-ranging data with the goal of being able to use the model on a multitude of ``downstream" tasks. 
The T5 model accepts an optional ``prefix" before the input to specify the task.
A sample input, as provided by \citeauthor{raffel2020exploring}, involves the translation of text from English to German. 
In this example, 
T5 would take the input as ``translate English to German: (text to be translate)." 
This flexibility allowed us to use our input and target emotions as the prefixes in  fine-tuning T5 for implementing the task of fine-grained emotional-paraphrasing.

\subsection{Paraphrasing} 
Paraphrasing is a key task in the NLG domain. It involves inputting text and returning one of similar meaning. 
Approaches to paraphrasing include the use of sentence pattern and feature statistics \citep{zhao2009application}, 
repurposing unsupervised style transfer \citep{krishna2020reformulating},
imitation and reinforcement learning \citep{du2019empirical},
a latent bag-of-words \citep{fu2019paraphrase},
and pointer-generator networks \citep{see2017get}.
Several models combine deep generative models with other modeling and training techniques:  
e.g., variations using reinforcement learning \citep{li2017paraphrase},
long short-term memory or LSTM \citep{gupta2018deep}, 
and stacked residual LSTM \citep{prakash2016neural}.
Other approaches focus on specific variations of paraphrasing. 
These include using reinforcement learning for the multi-paraphrase generation \citep{qian2019exploring}, weakly supervised learning for entailment-relation-aware paraphrasing \citep{sancheti2022entailment},
adversarial training for end-to-end paraphrasing \citep{yang2019end},
and syntax-guidance for guided paraphrasing \citep{kumar2020syntax}.

Transformer-based text-to-text models such as BART and T5 have become more popular for use in paraphrasing. 
Several studies have been done with the goal of improving these models' paraphrasing performance through joint paraphrase learning \citep{min2020advancing},
topic-focused training \citep{liu2019generating},
and combining copying and generative decoders \citep{cao2017joint}.
Work has also been done to combine Transformers and seq2seq models to improve paraphrase generation \citep{egonmwan2019transformer}

Equally important to paraphrasing is the evaluation of paraphrase quality. 
The most common metrics used for paraphrase evaluation are BLEU \citep{papineni2002bleu}, ROUGE \citep{lin2004rouge}, and METEOR \citep{banerjee2005meteor}. 
Recently, newer scores have also been proposed for evaluating text generation and paraphrase models include Google BLEU \citep{wu2016google} - a variation of the classic BLEU score, 
the BERTScore \citep{zhang2019bertscore} - obtained using the BERT model,  
MAUVE \citep{pillutla2021mauve}, BBScore \citep{shen2022revisiting}, and fine-grained pair-wise word interactions (PWI) scores \citep{he2016pairwise}.

\section{Fine-Grained \\ Emotional Paraphrasing}

To tackle the task of fine-grained emotional paraphrasing along emotion gradients, we propose a novel workflow as illustrated in Figure~\ref{fig:Approach1}. Given an input text, our workflow first identifies the current emotion of the text. Then it selects the desired target emotion for the paraphrase, per user guidance, utilizing an emotion transition graph that is based on emotion gradients. After that, it generates a prefix for the selected transition based on the source and target emotions. Finally, it sends the input text with the generated task prefix to our fine-tuned paraphrase generator to produce the target paraphrase. Below we discuss the four workflow components in detail. 

\subsection{Emotion Classifier}
The first step in our emotional paraphrasing workflow is to identify the emotion of the input text. 
This is done through our enhanced version of the GoEmotions model (see Section~\ref{SubSec:DatasetLabeling} for details of our enhancement). Given the input text, this classification model identifies the most compatible of the 28 emotions to feed into the transition graph. 
The GoEmotions model has a wider variety and more detailed array of emotions compared to emotion classifications such as Ekman's. This allows for more precise emotion classifications that enable 
fine-grained adjustment of paraphrase emotions. 

\begin{figure*}[t!]
    \centering
    \includegraphics[width=0.9\linewidth]{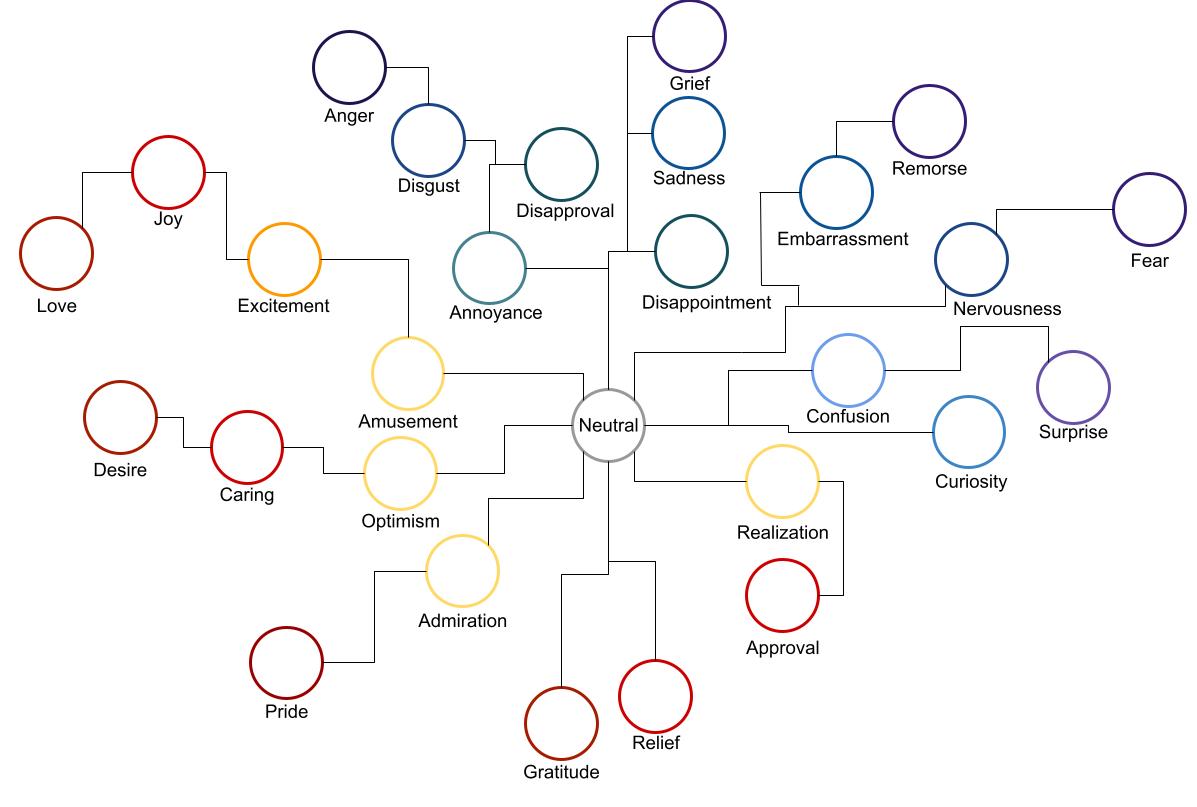}
    \caption{Proposed Emotion Transition Graph}
    \label{fig:TransitionGraph}
\end{figure*}

\subsection{Emotion Transition Graph}    
The second step in our workflow is user-guided emotional adjustment per the emotion transition graph as shown in Figure~\ref{fig:TransitionGraph}. This transition graph is based on the GoEmotions emotion heatmap created by \citeauthor{demszky2020goemotions}, which shows emotions as bridged by continuous gradients. 
Each group of emotions as shown in Table~\ref{EmotionClusters}, although close in sentiments, exhibits different levels of intensities.
We create this emotion transition graph by starting with the emotion clusters extrapolated from the GoEmotions emotion heatmap. At the center of the graph is the neutral emotion. Emotions from all other clusters can transition to the neutral emotion. Emotions within each cluster can transition to each other along their smooth emotion gradients. The edges between emotions within each cluster indicate an increase or decrease of emotional intensities and the emotions closer to neutral have lower intensities. We add the edges manually to reflect our understanding of emotional intensities. 
By following the transition graph, we can adjust emotion intensity.
For example, if the GoEmotions identifies the input emotion as anger, the transition graph could recommend a transition to annoyance, which we, as a user of the workflow, can adopt. 

\subsection{Prefix Generation}
The third step in our workflow is prefix generation. T5 features a multi-task design, i.e., many NLP tasks can be cast as text-to-text tasks and a prefix can be added to the input text to indicate the task at hand. The prefix generator utilizes this features and generates the prefix for the task of fined-grained emotional paraphrasing. Given the source emotion identified in the emotion classification step and the target emotion selected in the emotion transition step, the prefix is generated in the format of ``(input emotion) to (desired output emotion):" and it is placed in front of the input text. It guides the fine-tuned T5 model to paraphrase along the selected emotion gradient.  

\subsection{Paraphrase Generator}
The final step of our workflow is paraphrase generation which utilizes a fine-tuned T5 model to complete the task of fine-grained emotion paraphrasing along emotion gradients. The T5 model is fine-tuned with a dataset of paraphrase pairs that exemplify the emotion transitions along the emotional gradients on the transition graph. The fine-tuned model allows for precise emotional paraphrasing by inputting the emotion transition prefix and the input text, paraphrasing it, and outputting the resulting paraphrase that best fits the desired output emotion.

\section{Fine-Tuning \\ Text-to-Text Transformers}

In this section, we discuss how to fine-tune a text-to-text transformer model, namely T5, to produce our fine-grained paraphrase generator. 

\subsection{Task Formulation} 
Let \textit{i} be the input text and \textit{o} be the ID corresponding to the original emotion. 
Let \textit{t} be the ID corresponding to the target emotion, and \textit{p} be the emotional paraphrased output of \textit{i}. 
In the task of fine-grained emotional paraphrasing along emotion gradients, \textit{i}, \textit{o}, and \textit{t} are given to the T5 model in the text format: \textit{o to t: i}. 
The fine-tuned model will then output \textit{p}, 
a paraphrased version of \textit{i} where the underlying semantics of \textit{i} are kept and where the intensity of emotion is changed.

\subsection{Dataset Preparation}
\label{SubSec:DatasetPreparation}
The datasets that we used to fine-tune the T5 model were Google PAWS-Wiki  \citep{zhang2019paws}, Microsoft Research Paraphrase Corpus \citep{dolan2005automatically}, Quora Questions Pairs \citep{iyer2017qqpairs}, and Twitter Language-Net paraphrasing corpus \citep{lan2017continuously}. 

The Google PAWS (Paraphrase Adversaries from Word Scrambling) project produced multiple sets of paraphrasing input-output pairs. 
We chose to use to PAWS-Wiki Labeled (Final) data because they were generated by translation methods and human verified for accuracy.
The PAWS-Wiki corpus was pre-split into a training and a test sets. 
This allowed them to be readily inputted separately into two Python Pandas dataframes. 
In total, the PAWS-Wiki corpus provided 49,401 pairs of input-output training data and 8,000 testing pairs.

Similar to the PAWS-Wiki corpus, Microsoft Research Paraphrase Corpus (MRPC) was also pre-split into a training set and a test set. 
The MRPC was a compilation of human-annotated data from the news. 
Using a Pandas function, we concatenated the Microsoft training and test datasets to their respective Pandas dataframes. 
In total, 
the Microsoft corpus provided 5,801 pairs of input-output data.

The Quora corpus we used was the Quora Question Pairs (QQP). 
It was released with the goal of aiding the training of ``semantic equivalence" models,
similar to the goals of paraphrasing models, 
which suited our needs in fine-tuning for emotional paraphrasing. 
However, 
the Question Pairs dataset was not pre-split. 
Accordingly, 
we utilized the train\_test\_split function from the sklearn.model\_selection package to randomly select pairs from the QQP and split them into training and test sets.
In total, 
the QQP provided 111,948 training pairs and 37,317 test pairs. 
They were then appended to the Google and Microsoft data to create a combined training set consisting of 163,995 text pairs and a combined test set consisting of 46,399 pairs. 

The Twitter Language-Net corpus was a dataset created through associating tweets through URL. 
It consisted of 51,524 pairs of paraphrase candidates that were human labeled by six people to be paraphrases of each other. 
For the task of fine-grained emotional paraphrasing, 
we kept only the pairs with the majority of raters agreeing on its validity as a paraphrase pair. 
However, 
once the emotions were labeled for these pairs, 
the majority turned out have no dominant emotion or were labeled as \textit{neutral}. 
Once those were removed, 
the dataset only contained about 200 pairs. 
The Language-Net corpus also had an expanded set of 2,869,657 candidate pairs that had been scored using the Pair-wise Word Interactions (PWI) metric for evaluating paraphrasing. 
The PWI score ranges from 0 to 1, 
0 being the lowest quality, 
and 1 being an exact match. 
A variety of score thresholds such as 0.775 and 0.825 were tested to find the balance between quality and quantity when used for fine-tuning T5 in comparison with the combined dataset.

\subsection{Labeling Paraphrasing Datasets} 
\label{SubSec:DatasetLabeling}
The original GoEmotions model, for each input text, outputs a list of emotions that it identified as being ``possible" candidates for the emotion of the input text. It also outputs a ``score" that is a decimal from 0 to 1 representing the likelihood for that emotion to be the correct emotion. 
These scores are calculated by a Sigmoid function in the model. We instead modified the output code of the model to return the emotion, if any, that had a score over the set threshold of 0.5. Otherwise, the model would report no emotion. 

Before we were able to fine-tune T5 with the prepared datasets, 
we pre-processed both the training and test sets by removing any input-output pairs with blank emotions. 
In other words, 
the inputs for which the GoEmotions model could not identify an emotion with a high enough score to pass the threshold of 0.5 were removed. 
We also removed any pairs for which both emotions are identified as \textit{neutral} because they made up the large portion of the dataset and could skew our T5 fine-tuning results.
As a last step, any lines with matching input-output emotions were removed as they did not provide value in training the emotional paraphrasing models.   


\subsection{Fine-Tuning T5 \\ with Labeled Datasets}
We utilized the Simple Transformers package \citep{rajapakse2022simpletransformers} to perform T5 fine-tuning. 
It included efficient and streamlined implementation of several transformer models including BART and T5. 
From Simple Transformers, 
we pulled the T5 model and the training and prediction generation functions. 
We conducted the fine-tuning and evaluation on a desktop computer with an AMD Ryzen 7 5800x, 32GB RAM, and RTX 3080TI GPU. Due to a limited amount of GPU memory, 12GB precisely, we had to adopt a smaller batch size of 6. Each model was trained over 3 epochs during the fine-tuning process to achieve optimal results while avoiding over-fitting the model.

\section{Preliminary Evaluation}

\subsection{Evaluation Metrics}

We evaluated the emotional paraphrasing capabilities of our fine-tuned T5 models from two aspects: emotion transition and paraphrasing. To achieve this, we utilized a wide array of metrics from Hugging Face's evaluate package~\citep{huggingface:metrics}.

To evaluate the emotion-transition performance of the models, we utilized the {\em Exact Match} metric, which finds the percentage of the emotions of the generated paraphrases that match the target emotions. By comparing the emotions (as classified by GoEmotions) of the target text and the prediction output of each model, the Exact Match score indicated how well the model can perform the task of emotion-transition by revealing how many target emotions the model was able to successfully transition to.

To evaluate the paraphrasing capabilities of the models, we utilize a wide set of paraphrase metrics:
{\em BLEU}, 
{\em Google BLEU}, 
{\em ROUGE-1},
{\em ROUGE-2},
{\em ROUGE-L}, and 
{\em METEOR}. 
%
%
This set of metrics, all scored on a scale of 0 to 1, indicated the paraphrasing performance of each model by evaluating similarities of the target texts and model predictions.

Given an evaluation dataset of paraphrase pairs, we first labeled both the input and target texts with their emotion ID labels using GoEmotions. We then sent the following query to the fine-tuned model: ``(input emotion ID) to (target emotion ID): input text." We then labeled the prediction with an emotion and compute the Exact Match metrics using the target emotion and prediction emotions as well as the paraphrase metrics using the input text and the prediction. 


\subsection{Fine-Tuning Datasets}

Three datasets were used in fine-tuning and evaluation of the models as follows:
\begin{enumerate}
 \item The mixture dataset combined the Google PAWS-Wiki, Microsoft MRPC, and Quora QQP datasets and is referred to as \textbf{mix}.
 \item The Twitter dataset was created from the Twitter Language-Net corpus with the PWI metric set to 0.825. PWI thresholds of 0.775, 0.8, and 0.825 were applied to find desirable balance between the quantity and quality of the paraphrasing pairs, resulting in the selection of 0.825. The resulting dataset after applying this threshold is referred to as \textbf{twit0.825} 
 \item The \textbf{combined} dataset is created by merging the {\bf mix} and {\bf twit0.825} datasets. 
\end{enumerate}


\subsection{Regular vs. \\ Limited-Data Fine-Tuning}
As mentioned in Section~\ref{SubSec:DatasetPreparation}, 
each dataset was split into a larger \textit{training} set and a smaller \textit{test} set as follows. 
The two datasets were used in the regular and limited-data fine-tuning processes.
\begin{enumerate}
    \item \textbf{Regular Fine-Tuning}: Using the larger \textit{training} set to fine-tune the base T5 model and the smaller \textit{test} set to evaluate the fine-tuned model. This helps evaluate how well a fine-tuned model can perform with large amounts of training data. 
    \item \textbf{Limited-Data Fine-Tuning}: Swapping the \textit{training} and \textit{test} sets and using the smaller \textit{test} set to fine-tune the base T5 model and the larger \textit{training} to evaluate the model. This helps evaluate how well a fine-tuned model can perform with limited amounts of training data. 
\end{enumerate}
 
\subsection{Evaluations of Fine-tuned Models}
In both the regular and limited-data fine-tuning cases, three models were obtained by fine-tuning with the three emotion-labeled paraphrasing datasets, {\bf mix}, {\bf twit0.825}, and \textbf{combined}.   
During evaluation, a fourth model, the base T5 model (a.k.a., {\bf t5-base}) is used as a control model. 
It helped to demonstrate the improvements of the fine-tuned models. 
In the evaluation, each of the four models was applied to the three test datasets from {\bf mix}, {\bf twit0.825}, and \textbf{combined}.  


\subsection{Preliminary Results}

\begin{table}[t]
\caption{Results from Regular Fine-Tuning}
\centering
\footnotesize
\noindent
\renewcommand{\arraystretch}{1.2}
\begin{tabular}{@{} c l *{3}{c} @{}}
\hline
& & \multicolumn{3}{c@{}}{\textbf{Evaluation Sets}}\\
\textbf{Metrics} & \textbf{Models} & twit0.825 & mix &  combined \\
\hline
& twit0.825 & 0.377 & 0.213 & 0.223 \\
\textbf{Exact} & mix & 0.196 & 0.391 & 0.367 \\
\textbf{Match} & combined & 0.305 & 0.404 & 0.396 \\
& t5-base & 0.098 & 0.140 & 0.139 \\ 
\hline
& twit0.825 & 0.506 & 0.111 & 0.158 \\
& mix & 0.149 & 0.206 & 0.200 \\
\textbf{~ BLEU} & combined & 0.484 & 0.210 & 0.254 \\
& t5-base & 0.168 & 0.129 & 0.136 \\
\hline 
& twit0.825 & 0.469 & 0.148 & 0.182 \\
\textbf{Google} & mix & 0.169 & 0.235 & 0.228 \\
\textbf{BLEU} & combined & 0.448 & 0.241 & 0.271 \\
& t5-base & 0.173 & 0.161 & 0.163 \\
\hline
& twit0.825 & 0.735 & 0.483 & 0.507 \\
& mix & 0.467 & 0.519  & 0.514 \\
\textbf{~ROUGE-1} & combined & 0.717 & 0.524 & 0.545 \\
& t5-base & 0.462 & 0.440 & 0.443 \\
\hline
& twit0.825 & 0.637 & 0.253 & 0.291 \\
& mix & 0.279 & 0.280 & 0.279 \\
\textbf{~ROUGE-2} & combined & 0.613 & 0.286 & 0.318 \\
& t5-base & 0.352 & 0.228 & 0.241 \\
\hline
& twit0.825 & 0.728 & 0.441 & 0.468 \\
& mix & 0.435 & 0.483 & 0.479 \\
\textbf{~ROUGE-L} & combined & 0.711 & 0.490 & 0.514 \\
& t5-base & 0.445 & 0.406 & 0.411 \\
\hline
& twit0.825 & 0.709 & 0.517 & 0.536 \\
& mix & 0.416 & 0.499 & 0.491 \\
\textbf{~METEOR} & combined & 0.703 & 0.504 & 0.526 \\
& t5-base & 0.505 & 0.488 & 0.492 \\
\hline
\end{tabular}
\label{Tab:Regular-Fine-Tuning}
\end{table}

Table~\ref{Tab:Regular-Fine-Tuning} shows the evaluation results of regular fine-tuned models and Table~\ref{Tab:Limited-Data-Fine-Tuning} shows the results from limited-data fine-tuned models. Each table is organized by metrics. 
It can be observed from Table~\ref{Tab:Regular-Fine-Tuning} that in each metric for every dataset, the models trained using emotion-labeled paraphrase datasets outperformed the control t5-base model. This indicates fine-tuning with emotion-labeled paraphrases was able to help guide T5 in improving its ability at emotional paraphrasing. Therefore, our solution of fine-tuning T5 with emotion-labeled data is effective at the task of emotional paraphrasing. Another observation is that a model fine-tuned with a given training set outperforms on its corresponding test set when compared to other models that have not be exposed to the same data. For example, the model fine-tuned by twit0.825 has a better Exact Match score when applied to the twit0.825 evaluation set than does the fine-tuned mix model. This means that while the emotion-paraphrasing potential of fine-tuned models is strong, transfer learning still remains a challenge.

\begin{table}[t]
\caption{Results from Limited-Data Fine-Tuning}
\centering
\footnotesize
\noindent
\renewcommand{\arraystretch}{1.2}
\begin{tabular}{@{} c l *{3}{c} @{}}
\hline
& & \multicolumn{3}{c@{}}{\textbf{Evaluation Sets}}\\
\textbf{Metrics} & \textbf{Models} & twit0.825 & mix & combined \\
\hline
& twit0.825 & 0.240 & 0.178 & 0.184 \\
\textbf{Exact} & mix & 0.165 & 0.345 & 0.327 \\
\textbf{Match} & combined & 0.210 & 0.352 & 0.337 \\
& t5-base & 0.088 & 0.143 & 0.139 \\
\hline
& twit0.825 & 0.429 & 0.173 & 0.208 \\
& mix & 0.166 & 0.207 & 0.201 \\
\textbf{BLEU} & combined & 0.380 & 0.191 & 0.219 \\
& t5-base & 0.163 & 0.138 & 0.142 \\
\hline 
& twit0.825 & 0.394 & 0.198 & 0.222 \\
\textbf{Google}& mix & 0.169 & 0.241 & 0.231 \\
\textbf{BLEU} & combined & 0.359 & 0.228 & 0.244 \\
& t5-base & 0.166 & 0.169 & 0.168 \\
\hline
& twit0.825 & 0.702 & 0.538 & 0.554 \\
& mix & 0.477 & 0.532 & 0.525 \\
\textbf{ROUGE-1} & combined & 0.690 & 0.511 & 0.527 \\
& t5-base & 0.452 & 0.446 & 0.446 \\
\hline
& twit0.825 & 0.596 & 0.294 & 0.324 \\
& mix & 0.295 & 0.283 & 0.283 \\
\textbf{ROUGE-2} & combined & 0.568 & 0.264 & 0.291 \\
& t5-base & 0.337 & 0.233 & 0.243 \\
\hline
& twit0.825 & 0.695 & 0.497 & 0.517 \\
& mix & 0.450 & 0.497 & 0.490 \\
\textbf{ROUGE-L} & combined & 0.683 & 0.475 & 0.493 \\
& t5-base & 0.432 & 0.411 & 0.413 \\
\hline
& twit0.825 & 0.671 & 0.541 & 0.554 \\
& mix & 0.428 & 0.511 & 0.502 \\
\textbf{METEOR} & combined & 0.648 & 0.489 & 0.503 \\
& t5-base & 0.481 & 0.494 & 0.493 \\
\hline
\end{tabular}
\label{Tab:Limited-Data-Fine-Tuning}
\end{table}

While the datasets we utilized in our experiment were not sufficient for training a generalized model for the task of fine-grained emotional paraphrasing, the results of limited-data fine-tuning show that a generalized model may not be critical. As shown in Table~\ref{Tab:Limited-Data-Fine-Tuning}, the models were fine-tuned using the smaller \textit{test} sets meaning the models have been exposed to less data. The trend extrapolated from the results of limited-data fine-tuning suggests that even with less data, the fine-tuned models consistently outperform t5-base.
This is significant because it means that effective emotion-paraphrasing models for certain applications can be trained using small amounts of high quality data instead of utilizing a large dataset of generalized paraphrase pairs. Additional work is needed to verify this observation on other datasets. 

\begin{table}[t]
\caption{Results from Restricted Test Dataset}
\centering
\footnotesize
\noindent
\renewcommand{\arraystretch}{1.2}
\begin{tabular}{@{} c l *{3}{c} @{}}
\hline
& & \multicolumn{3}{c@{}}{\textbf{Evaluation Sets}}\\
\textbf{Metrics} & \textbf{Models} & twit0.825 & mix &  combined \\
\hline
& twit0.825 & 0.489 & 0.141 & 0.347 \\
\textbf{Exact} & mix & 0.100 & 0.126 & 0.093 \\
\textbf{Match} & combined & 0.467 & 0.274 & 0.342 \\
& t5-base & 0.133 & 0.111 & 0.160 \\
\hline
& twit0.825 & 0.503 & 0.327 & 0.406 \\
& mix & 0.211 & 0.343 & 0.283 \\
\textbf{BLEU} & combined & 0.495 & 0.340 & 0.400 \\
& t5-base & 0.180 & 0.215 & 0.186 \\
\hline 
& twit0.825 & 0.488 & 0.371 & 0.411 \\
\textbf{Google} & mix & 0.227 & 0.386 & 0.315 \\
\textbf{BLEU} & combined & 0.477 & 0.382 & 0.409 \\
& t5-base & 0.190 & 0.256 & 0.212 \\
\hline
& twit0.825 & 0.78 & 0.755 & 0.757 \\
& mix & 0.516 & 0.721 & 0.638 \\
\textbf{ROUGE-1} & combined & 0.764 & 0.732 & 0.729 \\
& t5-base & 0.511 & 0.612 & 0.550 \\
\hline
& twit0.825 & 0.643 & 0.496 & 0.559 \\
& mix & 0.337 & 0.441 & 0.403 \\
\textbf{ROUGE-2} & combined & 0.626 & 0.455 & 0.515 \\
& t5-base & 0.387 & 0.367 & 0.364 \\
\hline
& twit0.825 & 0.767 & 0.638 & 0.685 \\
& mix & 0.488 & 0.604 & 0.562 \\
\textbf{ROUGE-L} & combined & 0.751 & 0.637 & 0.670 \\
& t5-base & 0.493 & 0.507 & 0.485 \\
\hline
& twit0.825 & 0.733 & 0.712 & 0.720 \\
& mix & 0.456 & 0.661 & 0.588 \\
\textbf{METEOR} & combined & 0.728 & 0.693 & 0.691 \\
& t5-base & 0.558 & 0.591 & 0.568 \\
\hline
\end{tabular}
\label{Tab:Regular-Fine-Tuning-TG}
\end{table}

\begin{table}[t]
\caption{Results from Restricted Training and Test}
\centering
\footnotesize
\noindent
\renewcommand{\arraystretch}{1.2}
\begin{tabular}{@{} c l *{3}{c} @{}}
\hline
& & \multicolumn{3}{c@{}}{\textbf{Evaluation Sets}}\\
\textbf{Metrics} & \textbf{Models} & twit0.825 & mix & combined \\
\hline
& twit0.825 & 0.5 & 0.03 & 0.227 \\
\textbf{Exact} & mix & 0.122 & 0.274 & 0.204 \\
\textbf{Match} & combined & 0.556 & 0.289 & 0.391 \\
& t5-base & 0.078 & 0.104 & 0.164 \\
\hline
& twit0.825 & 0.543 & 0.324 & 0.432 \\
& mix & 0.307 & 0.382 & 0.351 \\
\textbf{BLEU} & combined & 0.508 & 0.374 & 0.431 \\
& t5-base & 0.192 & 0.193 & 0.194 \\
\hline 
& twit0.825 & 0.541 & 0.378 & 0.432 \\
\textbf{Google} & mix & 0.319 & 0.421 & 0.377 \\
\textbf{BLEU} & combined & 0.495 & 0.417 & 0.444 \\
& t5-base & 0.204 & 0.227 & 0.221 \\
\hline
& twit0.825 & 0.802 & 0.761 & 0.779 \\
& mix & 0.654 & 0.766 & 0.728 \\
\textbf{ROUGE-1} & combined & 0.786 & 0.758 & 0.765 \\
& t5-base & 0.534 & 0.558 & 0.558 \\
\hline
& twit0.825 & 0.711 & 0.466 & 0.568 \\
& mix & 0.506 & 0.502 & 0.501 \\
\textbf{ROUGE-2} & combined & 0.671 & 0.484 & 0.547 \\
& t5-base & 0.42 & 0.335 & 0.376 \\
\hline
& twit0.825 & 0.796 & 0.621 & 0.692 \\
& mix & 0.646 & 0.65 & 0.652 \\
\textbf{ROUGE-L} & combined & 0.777 & 0.64 & 0.685 \\
& t5-base & 0.52 & 0.474 & 0.494 \\
\hline
& twit0.825 & 0.763 & 0.669 & 0.71 \\
& mix & 0.608 & 0.688 & 0.666 \\
\textbf{METEOR} & combined & 0.765 & 0.705 & 0.728 \\
& t5-base & 0.588 & 0.556 & 0.573 \\
\hline
\end{tabular}
\label{Tab:Restricted-Fine-Tuning-TG}
\end{table}

\subsection{Performance on Transition Graph} 

In the fine-tuning evaluations, as shown in Table~\ref{Tab:Regular-Fine-Tuning} and Table~\ref{Tab:Limited-Data-Fine-Tuning},
we did not restrict the training and test datasets to only the emotion transitions that are present on the transition 
graph as shown in Figure~\ref{fig:TransitionGraph}. In this section, we explore whether restricting the datasets as such
has significant effects on the emotional paraphrasing capabilities of the fine-tuned models. Table~\ref{Tab:Regular-Fine-Tuning-TG} 
recalculates all the metrics without retraining the models and only by restricting the test datasets according to
the transition graph. It can be observed that the fine-tuned model has slightly better performance on the restricted test 
datasets compared with the original general test datasets. Table~\ref{Tab:Restricted-Fine-Tuning-TG} shows the results from 
fine-tuning and evaluating the models on training and test datasets that were both restricted according to the 
transition graph. It can be observed that the performance, especially for emotion-transitions, is further improved over the unrestricted case. Our hypothesis is that emotion transitions following emotion gradients captured by the transition graph better align with the paraphrasing knowledge already embedded in the underlying T5 model. We will further validate this hypothesis in future studies.

\section{Related Work} 

Emotional paraphrasing is a task that has been closely studied.
In \citep{casas2021emotional}, GPT models were utilized for 
emotional paraphrasing. Input text was paraphrased to fit 
within one of Ekman's six emotional categories. Six fine-tuned 
GPT models (one for each emotion) are utilized to ``enhance" the 
emotion of the input text which was then paraphrased. While 
our task shares the similar goal in emotional paraphrasing, 
our fine-grained paraphrasing solution is more flexible. It utilizes a
more fine-grained emotion categorization, conducts
emotional transition based on the emotion of the input text,  
and is capable of transitioning to various emotions along 
emotion gradients on the transition graph.  


As our task lowers emotional intensity of input texts, and 
thereby lowering the strong psychological effects that intense 
emotional interactions can bring, it relates to the task 
of positive reframing \citep{ziems2022inducing}. Both focus on 
altering the emotions of texts, while preserving its 
underlying connotations. However, the task of positive 
reframing emphasized altering the input text into a positive 
emotion while our task does not translate every emotion into 
a positive one, but rather lowers the intensity of emotions, 
which allows negative and positive emotions alike.

Our goal of lowering the intensity of emotions in text is 
related to, but different from the task of neutralizing bias \citep{pryzant2020automatically}. 
Neutralizing bias strives to eliminate all bias, which 
results in most paraphrased text being classified as \textit{neutral}. 
In our task of emotion paraphrasing along emotion-gradients, we aim 
to preserve the base meaning and tone while lowering the intensity 
of the input text. In that way, the text still expresses its original view or 
belief,  but in a less provocative or intense manner.

\section{\hspace{-0.02in}Conclusions and Future Work}

In this paper, we introduced a new task of fine-grained emotional 
paraphrasing along emotion gradients. This task seeks to adjust
the emotional intensities of the input texts following smooth 
variations in affective dimensions while preserving the meanings 
of the inputs. We developed a framework for addressing this task 
through fine-tuning text-to-text Transformer models through multi-task 
learning. Our implementation of this framework on T5 has demonstrated 
that fine-tuned T5 models by fine-grained emotional paraphrasing data
perform significantly better than the baseline T5 model on test 
data in both emotion transition and paraphrasing. 


For future work, we will explore the following two directions.  
First, emotional classification plays a key role in emotional paraphrasing. 
A less accurate or even incorrect classification of input and output
text emotions will guide paraphrasing the wrong way and induce 
incorrect decisions on how to conduct emotion transition. We will explore
improvement to the GoEmotions model or explore additional models for 
emotion classification. Second, there is still large room of improvement
for emotional paraphrasing. We will pursue better datasets for emotional
fine-tuning or even develop new datasets for this purpose. We will also
investigate more customized models beyond the baseline text-to-text models. 
For evaluation, we plan to conduct human studies when appropriate. 


\bibliographystyle{apacite}
\bibliography{FGEPAEG.bib}

\end{document}